\documentclass{article} 
\usepackage{nips14submit_e,times}
\usepackage{url}

\usepackage[]{algorithm2e}

\usepackage{amsmath, amsthm, amssymb}

\usepackage{multirow}
\newcommand{\comment}[1]{\textcolor{red}{#1}}
\usepackage{rotating}
\usepackage{tabu}
\long\def\comment#1{}

\newcommand{\refSection}[1]{Section~\ref{#1}}
\newcommand{\refTable}[1]{Table~\ref{#1}} 
\newcommand{\refFigure}[1]{Figure~\ref{#1}}

\newcommand{\refCite}[1]{\cite{#1}}
\newcommand{\refCitep}[1]{\cite{#1}}

\newcommand{\refEq}[1]{eq~\ref{#1}}

\newcommand{\ie}[0]{\textit{i.e.},~}
\newcommand{\eg}[0]{\textit{e.g.},~}
\newcommand{\aka}[0]{\textit{a.k.a}.~}
\newcommand{\etal}{\textit{et al.}}

\newcommand\scalemath[2]{\scalebox{#1}{\mbox{\ensuremath{\displaystyle #2}}}}
\newcommand{\x}[0]{\ensuremath{x}}
\newcommand{\xh}[0]{\ensuremath{\widehat{x}}}
\newcommand{\xv}[0]{\ensuremath{\mathbf{x}}}
\newcommand{\xvh}[0]{\ensuremath{\widehat{\mathbf{x}}}}
\newcommand{\dd}[2]{\ensuremath{d_{#1 \rightarrow #2}}}
\newcommand{\ddd}[2]{\ensuremath{p_{#1 < #2}}}
\newcommand{\set}[1]{\ensuremath{\mathcal{#1}}}
\newcommand{\f}{\ensuremath{f}}

\newcommand{\I}[0]{\mathbb{I}_{\infty}}

\newcommand{\back}[1]{\ensuremath{\setminus #1}}
\newcommand{\p}{\ensuremath{\mu}}

\newcommand{\ph}{\ensuremath{\widehat{\mu}}}

\newcommand{\Ms}[2]{\ensuremath{\nu_{#1 \to #2}}}
\newcommand{\Mst}[2]{\ensuremath{\widetilde{\nu}_{#1 \to #2}}}
\newcommand{\G}[0]{\ensuremath{\set{G}}}
\newcommand{\V}[0]{\ensuremath{\set{V}}}
\newcommand{\E}[0]{\ensuremath{\set{E}}}
\newcommand{\C}[0]{\ensuremath{\set{C}}}
\newcommand{\s}[0]{\ensuremath{\set{S}}}
\newcommand{\Is}[0]{\ensuremath{\set{I}}}

\title{Augmentative Message Passing for Traveling Salesman Problem and Graph Partitioning}

\author{
Siamak Ravanbakhsh \\
Department of Computing Science\\
University of Alberta\\
Edmonton, AB T6G 2E8 \\
\texttt{mravanba@ualberta.ca} \\
\And
Reihaneh Rabbany \\
Department of Computing Science \\
Edmonton, AB T6G 2E8 \\
\texttt{rabbanyk@ualberta.ca} \\
\AND
Russell Greiner \\
Department of Computing Science\\
University of Alberta\\
Edmonton, AB T6G 2E8 \\
\texttt{rgreiner@ualberta.ca} \\
}

\nipsfinalcopy 

\begin{document}

\maketitle

\begin{abstract}
The cutting plane method is an augmentative constrained optimization procedure that
is often used with continuous-domain optimization techniques such as linear 
and convex programs. We investigate the viability of a similar idea within message passing -- which produces integral solutions – in the context of two combinatorial problems: 1) For Traveling Salesman Problem (TSP), we propose
a factor-graph based on Held-Karp formulation, with an exponential number of
constraint factors, each of which has an exponential but sparse tabular form. 2)
For graph-partitioning (\aka community mining) using modularity optimization,
we introduce a binary variable model with a large number of constraints that enforce formation of cliques. In both cases we are able to derive surprisingly simple
message updates that lead to competitive solutions on benchmark instances. In particular for TSP
we are able to find near-optimal solutions in the time that empirically grows with $N^3$, demonstrating that augmentation is practical and efficient.
\end{abstract}

\section{Introduction} 
Probabilistic Graphical Models (PGMs) provide a principled approach to 
approximate constraint optimization for NP-hard problems.
This involves a message passing procedure (such as max-product Belief Propagation; BP) 
to find an approximation to {\em maximum a posteriori} (MAP) solution.
Message passing methods are also attractive 
as they are easily mass parallelize. 
This has contributed to their application in approximating many NP-hard problems,
including constraint satisfaction \refCite{mezard_analytic_2002,ravanbakhsh2014},
constrained optimization \refCite{frey2007clustering,bayati2005maximum},
min-max optimization \refCite{ravanbakhshmin},
and integration \refCite{huang2009approximating}.

The applicability of PGMs to discrete optimization problems is limited by the size and number of 
factors in the factor-graph. While many recent attempts have been made to reduce the complexity of
message passing over high-order factors \refCite{potetz2008efficient,gupta2007efficient,tarlow2010hop}, 
to our knowledge no published result addresses the issues of dealing with large number of factors. 
We consider a scenario where a large number of factors represent hard constraints 
and ask \emph{whether it is possible to find a feasible solution by considering
only a small fraction of these constraints.}

The idea is to start from a PGM corresponding to a tractible subsset of constraints, and after obtaining an approximate MAP solution using min-sum BP,
augment the PGM with the set of constraints that are violated in the current solution. This general idea has been extensively 
studied under the term \emph{cutting plane methods} in different settings. 
Dantzig \etal\ \refCitep{dantzig1954solution} first investigated this idea in the context 
of TSP and Gomory \etal \refCitep{gomory1958outline} provided a elegant method to generate violated 
constraints in the context of finding integral solutions to linear programs (LP). 
It has since been used to also solve a variety of nonlinear optimization problems.

The requirements of the cutting plane method are 1) availability
of an optimal solver;  often an LP solver,  2) a procedure to obtain violated
constraints and 3) operating in real domain $\Re^d$; hence the term ``plane''. 
Recent studies show that message passing -- which finds integral solutions -- can be much faster than LP in finding approximate MAP assignments for structured 
optimization problems~\refCite{yanover2006linear}.
This further motivates our inquiry regarding the viability of augmentation for message passing.
We present an affirmative answer to this question in application to two combinatorial problems. 
\refSection{sec:pgm} introduces our factor-graph formulations for Traveling Salesman Problem (TSP) and graph-partitioning.
\refSection{sec:BP} derives simple message update equations for these
factor-graphs and reviews our augmentation scheme.  
Finally, \refSection{sec:experiments} presents experimental results for both applications.

\section{Background and Representation}\label{sec:pgm} 
Let $x = \{x_1,\ldots,x_D\} \in \set{X} = \set{X}_1 \times \set{X}_2\ldots \times \set{X}_D$ denote an instance
of a tuple of discrete variables. 
Let $x_\Is$ refer to a sub-tuple, 
where $\Is \subseteq \{ 1,\ldots, D\}$ indexes a subset of these variables.
Define the energy function 
$\f(\x) \; \triangleq \; \sum_{\Is \in \set{F}} \f_\Is(\x_\Is)$ 
where $\set{F}$ denotes the set of factors.
Here the goal of inference is to find an assignment with minimum energy
$x^* \; = \; \arg_x\min \f(x)$.
This model can be conveniently represented using a bipartite graph, known as factor-graph \refCite{kschischang_factor_2001},
where a factor node $\f_{\Is}(x_{\Is})$ is connected to a variable node $\x_i$\ \textit{iff}\ $i \in \Is$.

\subsection{Traveling Salesman Problem} 
\label{sec:TSP}
A Traveling Salesman Problem (TSP) seeks the minimum length tour of $N$ cities that visits each city exactly once. 
TSP is $\set{NP}$-hard,
and for general distances, no constant factor approximation to this problem is possible \refCite{papadimitriou1977euclidean}. 
The best known exact solver,
due to Held \etal \refCitep{held_dynamic_1962}, uses dynamic programming to reduce the cost of enumerating all orderings from
$\set{O}(N!)$ to $\set{O}(N^2 2^N)$.
The development of many (now) standard optimization techniques,
such as simulated annealing, 
mixed integer linear programming, 
dynamic programming, 
and ant colony optimization 
are closely linked with advances in solving TSP.
Since Dantzig \etal \refCitep{dantzig1954solution} manually applied the cutting plane method to 49-city problem, 
a combination of more sophisticated cuts, used with branch-and-bound techniques~\refCite{padberg1991branch}, 
has produced the state-of-the-art TSP-solver, Concorde \refCite{concorde}. 
Other notable results on very large instances have been reported by Lin–Kernighan heuristic \refCite{lkh} 
that continuously improves a solution by exchanging nodes in the tour. 
In a related work, Wang \etal \refCite{wangmessage} 
proposed a message passing solution to TSP.
However their method does not scale beyond small toy problems (authors experimented with $N=5$ cities).  
For a readable historical background of the state-of-the-art in TSP and its various applications, see \refCitep{applegate2006traveling}.

\subsubsection{TSP Factor-Graph} 
Let $\G = (\V, \E)$ denote a graph, where $\V = \{v_1,\ldots,v_N\}$ is the set of nodes and 
the set of edges $\E$ contains 
$e_{i-j}$ \textit{iff} $v_i$ and $v_j$ are connected. 
Let $x = \{ x_{e_1},\ldots,x_{e_M}\} \in \set{X} = \{0,1\}^{M}$ be a set of binary variables, one for each edge in the graph (\ie $M = |\E|$) where 
 we will set
$x_{e_m} = 1$ \textit{iff} $e_m$ is in the tour.
For each node $v_i$, let $\partial v_i = \{e_{i-j}\ | \ e_{i-j} \in \E\}$ denote the edges adjacent to $v_i$.
Given a distance function $d: \E \to \Re$, define the  \textbf{local factors} for each edge $e \in \E$ as
$\f_{e}(x_e)\ =\  x_e\ d(e)$ -- so this is
either $d(e)$ or zero. 
Any valid tour satisfies the following necessary and sufficient constraints -- \aka Held-Karp constraints \refCite{held1970traveling}:
\\[1ex]
\textbf{1. Degree constraints:} Exactly two edges that are adjacent to each vertex should be in the tour. 
Define the factor
$\f_{\partial v_i}(x_{\partial v_i}): \{0,1\}^{|\partial v_i|} \to \{0,\infty\}$ 
to enforce this constraint 
$$ 
\f_{\partial v_i}(x_{\partial v_i})\  \triangleq\ \I\left(\sum_{e \in \partial v_i} x_e = 2\right)\quad\quad \forall v_i \in \V
$$ 
where $\I(condition) \triangleq 0$ \textit{iff} the condition is satisfied and $+\infty$ otherwise.

\medskip
\textbf{2. Subtour constraints:} Ensure that there are no short-circuits -- 
\ie there are no loops that contain strict subsets of nodes. 
To enforce this, for each 
 $\s \subset \V$, define 
$\delta(\s)\; \triangleq\; \{e_{i-j} \in \E \ | \ v_i \in \s, v_j \notin \s \}$
to be the set of edges, with one end in $\s$ and the other end in $\V \back \s$.

We need to have at least two edges leaving each subset $\s$. 
The following set of factors enforce these constraints
$$ 
\f_{\delta(\s)}(x_{\delta(\s)})\ =\ \I\left(\sum_{x_e \in \s} x_e \geq 2\right)\quad \quad 
\forall\, \s \subset \V ,\ \  \s \neq \emptyset
$$ 

\smallskip
These three types of factors define a factor-graph, 
whose minimum energy configuration 
is the smallest tour for TSP.

\subsection{Graph Partitioning} 
Graph partitioning --\aka community mining-- 
is an active field of research 
that has recently produced a variety of community detection methods
(\eg see \refCite{leskovec2010empirical} and its references), 
a notable one of which is Modularity maximization~\refCite{Newman04}. 
However, exact optimization of Modularity is $\set{NP}$-hard \refCite{brandes2008modularity}.
Modularity is closely related to fully connected Potts graphical models \refCite{reichardt2004detecting}. 
However, due to full connectivity of PGM, message passing is not able to find good solutions.
Many have proposed various other heuristics for modularity optimization 
\refCite{Clauset05RModularity,newman2006finding,reichardt2004detecting,
ronhovde2010local,blondel2008fast}.
We introduce a factor-graph representation of this problem 
 that has a large number of factors.
We then discuss a stochastic but sparse variation of modularity that enables us to efficiently partition relatively large sparse graphs.

\subsubsection{Clustering Factor-Graph} 
Let $\G = (\V, \E)$ be a graph, with a weight function $\widetilde{\omega}: \V \times \V \to \Re$,
where $\widetilde{\omega}(v_i,v_j) \neq 0$  \textit{iff} $e_{i:j} \in \E$.
Let $Z \;=\; \sum_{v_1,v_2 \in \V} \widetilde{\omega}(v_1, v_2)$ and $\omega(v_i, v_j) \triangleq \frac{\widetilde{\omega}}{2 Z}$ be the normalized weights. Also let 
$\omega(\partial v_i) \triangleq \sum_{v_j} \omega(v_i, v_j)$ denote the normalized degree of node $v_i$.
Graph clustering using modularity optimization
seeks a partitioning of the nodes into unspecified number of clusters $\set{C} = \{\set{C}_1,\ldots,\set{C}_K\}$, maximizing
\begin{align}\label{eq:modularity}
q(\C) \; =\; \sum_{\C_i \in \C} \;\;\sum_{v_i,v_j \in \C_i} \bigg ( \omega(v_i, v_j)\ -\ \omega(\partial v_i)\, \omega(\partial v_j) \bigg )
\end{align}
The first term of modularity is proportional to within-cluster
edge-weights. The second term is proportional to the expected number of within cluster edge-weights for a null model with the same weighted node degrees for each node $v_i$. 

Here the null model is a fully-connected graph. 
We generate a random \textit{sparse null model} with $M_{null} < \alpha M$ weighted edges ($\E_{null}$), by randomly sampling two nodes, each drawn independently from $\mathbb{P}(v_i) \propto \sqrt{\omega(\partial v_i)}$, and connecting them with a 
weight proportional to $\widetilde{\omega}_{null}(v_i, v_j) \propto \sqrt{\omega(\partial v_i)\omega(\partial v_j)}$. If they have been already
connected, this weight is added to their current weight.
We repeat this process $\alpha M$ times, however since some of the edges are repeated, the total number of edges in the null model may be under $\alpha M$.
Finally the normalized edge-weight in the sparse null model is
$
\omega_{null}(v_i, v_j) \triangleq \frac{\widetilde{\omega}_{null}(v_i, v_j)}{2  \sum_{v_i,v_j} \widetilde{\omega}_{null}(v_i, v_j)}.
$
It is easy to see that this generative process in expectation produces the fully connected null model.\footnote{The choice of using square root of 
weighted degrees for both sampling and weighting is to reduce the variance. One may also use pure importance sampling (\ie use
the product of weighted degrees for sampling and set the edge-weights in the null model uniformly), or uniform sampling of edges, where the edge-weights of
the null model are set to the product of weighted degrees.} 


Here we use the following binary-valued factor-graph formulation. 
Let $x = \{ x_{i_1:j_1},\ldots,x_{i_L:j_L}\} = \{0,1\}^{L}$ be a set of binary variables, 
one for each edge $e_{i:j} \in \E \cup \E_{null}$ -- \ie $|\E \cup \E_{null}| = L$.
Define the \textbf{local factor} for each variable as 
$\f_{i:j}(x_{i:j}) = - x_{i-j} (\omega(v_i,v_j) - \omega_{null}(v_i, v_j))$.
The idea is to enforce formation of cliques, while minimizing the sum of local factors.
By doing so the negative sum of local factors evaluates to  modularity (\refEq{eq:modularity}).
For each three edges $e_{i:j},e_{j:k},e_{i:k} \in \E \cup \E_{null}, i < j< k$ that form a triangle, define a \textbf{clique constraint} as 
$$
\f_{\{i:j, j:k, i:k\}}(x_{i:j}, x_{j:k}, x_{i:k}) \triangleq \\
\I(x_{i:j} + x_{j:k} + x_{i:k} \neq 2)
$$ 
These factors ensure the formation of cliques -- 
\ie if the weights of two edges that are adjacent to the same node are non-zero, the third edge in the triangle should also have non-zero weight.
The computational challenge here is the large number of clique constraints.
Brandes \etal \refCite{brandes2008modularity} use a similar LP formulation. 
However, since they include
all the constraints from the beginning and the null model is fully connected, their method is only applied to small toy problems.

\section{Message Passing}\label{sec:BP} 
Min-sum belief propagation is an inference procedure, in which
a set of messages are exchanged between variables and factors.
The factor-to-variable ($\Ms{\Is}{e}$) and variable-to-factor ($\Ms{e}{\Is}$) messages are defined as 
\begin{align}
\Ms{e}{\Is}(\x_e) \quad &\triangleq \quad \sum_{\Is' \ni e, \Is' \neq \Is } \Ms{\Is'}{e}(x_e) \label{eq:miI}\\
\Ms{\Is}{e}(\x_e)  \quad &\triangleq  \quad \min
    \bigg\{ \f_{\Is}(x_{\Is \back e}, x_e) \sum_{e' \in \Is \back e} \Ms{e'}{\Is}(x_{e'}) \bigg\}_{x_{\Is \back e}}
\label{eq:mIi}
\end{align}
where $\Is \ni e$ indexes all factors that are adjacent to the variable $x_e$ on the factor-graph.
Starting from an initial set of messages, this recursive update is performed until convergence.

This procedure is exact on trees, factor-graphs with single cycle 
as well as some special settings \refCite{bayati2005maximum}. 
However it is found to produce 
good 
approximations in general loopy graphs.
When BP is exact, the set of local beliefs 
$\p_e(\x_e)\; \triangleq \; \sum_{\Is \ni e} \Ms{\Is}{e}(x_e)$
indicate the minimum value that can be obtained for a particular assignment of $x_e$.
When there are no ties, the joint assignment $x^*$, 
obtained by minimizing individual local beliefs, is optimal.

When BP is not exact or the marginal beliefs are tied, a \textbf{decimation} procedure can improve 
the quality of final assignment. Decimation involves fixing a subset of variables to their most biased
values, and repeating the BP update. This process is repeated until all variables are fixed.

Another way to improve performance of BP when applied to loopy graphs is to use \textbf{damping}, which
often prevents oscillations:
$\Ms{\Is}{e}(\x_e)\; = \; \lambda \Mst{\Is}{e}(\x_e) + (1-\lambda)\Ms{\Is}{e}(\x_e)$.
Here $\Mst{\Is}{e}$ is the new message as calculated by \refEq{eq:mIi}
and $\lambda \in (0,1]$ is the damping parameter. Damping can also be applied
to variable-to-factor messages. 

When applying BP equations eqs~\ref{eq:miI}, \ref{eq:mIi} 
to the TSP and clustering factor-graphs,
as defined above,
we face two computational challenges: 
(a) Degree constraints for TSP can depend on $N$ variables, resulting in
$\set{O}(2^N)$ time complexity of calculating factor-to-variable messages. For 
subtour constraints, this is even more expensive as $\f_{\s}(x_{\delta(\s)})$ depends on $\set{O}(M)$ (recall $M = |\E|$ which can be $\set{O}(N^2)$) variables.
(b) The complete TSP factor-graph has $\set{O}(2^{N})$ subtour constraints. Similarly the clustering factor-graph
can contain a large number of clique constraints. 
For the fully connected null model, we need $\set{O}(N^3)$ such factors and 
even using
the sparse null model -- assuming a random edge probability \aka Erdos-Reny graph -- there are $\set{O}(\frac{L^3}{N^6} N^3) = \set{O}(\frac{L^3}{N^3})$
triangles in the graph (recall that $L = |\E \cup \E_{null} |$).
   In the next section, we derive the compact form of BP messages for both problems. 
In the case of TSP, we show 
how to exploit 
the sparsity of degree and subtour constraints
to 
calculate the factor-to-variable messages in 
 $\set{O}(N)$ and $\set{O}(M)$ respectively.


\subsection{Closed Form of Messages} 
For simplicity we work with
normalized message $\Ms{\Is}{e} \triangleq \Ms{\Is}{e}(1) - \Ms{\Is}{e}(0)$,
which is equivalent to assuming  $\Ms{\Is}{e}(0) = 0\  \forall \Is, e$.
The same notation is used for variable-to-factor message, and marginal belief. 
We refer to the normalized marginal belief, $\p_e = \p_e(1) - \p(0)_e $ as bias.

Despite their exponentially large tabular form, both degree and subtour constraint factors for TSP are sparse.
Similar forms of factors is studied in several previous works \refCite{potetz2008efficient, gupta2007efficient,tarlow2010hop}.
By calculating the closed form of these messages for \textbf{TSP factor-graph}, we observe that they have a surprisingly simple form.
Rewriting \refEq{eq:mIi}\ for \textit{degree constraint} factors, we get:
\begin{align}
  \Ms{\partial v_i}{e}(1)  =  \min \{ \Ms{e'}{\partial v_i} \}_{e' \in \partial v_i \back e}
\quad
,
\quad
  \Ms{\partial v_i}{e}(0)  =  \min \{ \Ms{e'}{\partial v_i} + \Ms{e''}{\partial v_i} \}_{e',e'' \in \partial v_i \back e}\label{eq:mIi_degree}
\end{align}
where we have dropped the summation and the factor from \refEq{eq:mIi}. 
For $x_e = 1$,
in order to have $\f_{\partial v_i}(x_{\partial i}) < \infty$, only \textit{one} other $x_{e'} \in x_{\partial v_i}$ should be non-zero. 
On the other hand, we know that messages are normalized
such that $\Ms{e}{\partial v_i}(0) = 0\ \  \forall v_i, e \in \partial v_i$,
which means they can be ignored in the summation. 
For $x_e = 0$, in order to satisfy the constraint factor, \textit{two} of the adjacent variables should have a non-zero value. 
Therefore we seek
two such incoming messages with minimum values. 
Let $\min[k] \set{A}$ denote the $k^{th}$ smallest value in the set $\set{A}$ -- \ie 
$\min \set{A} \equiv \min[1] \set{A}$.
We combine the updates above 
to get a ``normalized message'', $\Ms{\partial v_i}{e}$, which is simply the negative of the second largest incoming message (excluding $\Ms{e}{\partial v_i}$) 
to the factor $\f_{\partial v_i}$: 
\begin{align}
\Ms{\partial v_i}{e} = \Ms{\partial v_i}{e}(1) - \Ms{\partial v_i}{e}(0) 
= - \min[2] \{\Ms{e'}{\partial v_i}\}_{e' \in \partial v_i \back e} \label{eq:mIi_degree}
\end{align}

Following a similar procedure, factor-to-variable messages for \textit{subtour constraints} is given by
\begin{align}
  \Ms{\delta(\s)}{e} = -\max \{0 , \min[2] \{\Ms{e'}{\delta(\s)}\}_{e' \in \delta(\s) \back e}\} \}\label{eq:mIi_subtour}
\end{align}
Here while we are searching for the minimum incoming message, if we encounter two messages with negative or zero values, we can safely assume $\Ms{\delta(\s)}{e} = 0$, 
and stop the search. This results in significant speedup in practice. 
Note that both \refEq{eq:mIi_degree} and \refEq{eq:mIi_subtour} only need to calculate the second smallest message in the set $\{\Ms{e'}{\delta(\s)}\}_{e' \in \delta(\s) \back e}$.
In the asynchronous calculation of messages, this minimization should be repeated for each outgoing message. However in a \textbf{synchronous update} by finding three smallest incoming messages to each factor,
we can calculate all the factor-to-variable messages at the same time.

For the \textbf{clustering factor-graph}, the clique factor is satisfied only if either zero, one, or all three of the variables in its domain are non-zero.
The factor-to-variable messages are given by
\begin{align}
 &\Ms{\{i:j, j:k, i:k\}}{i:j}(0) = 
\min \{0 ,\; \Ms{j:k}{\{i:j, j:k, i:k\}} ,\; \Ms{i:k}{\{i:j, j:k, i:k\}}\} \notag\\
&\Ms{\{i:j, j:k, i:k\}}{i:j}(1) = 
\min \{0 ,\; \Ms{j:k}{\{i:j, j:k, i:k\}} \;+\; \Ms{i:k}{\{i:j, j:k, i:k\}}\}\label{eq:mIi_clustering_1}
\end{align}
For $x_{i:j} = 0$, the minimization is over three feasible cases
(a)~$x_{j:k} = x_{i:k} = 0$,
(b)~$x_{j:k} = 1, x_{i:k} = 0$ and 
(c)~$x_{j:k} = 0, x_{i:k} = 1$.
For $x_{i:j} = 1$, there are two feasible cases (a)~$x_{j:k} = x_{i:k} = 0$
and (b)~$x_{j:k} = x_{i:k} = 1$. 
Normalizing these messages we have
\begin{align}\label{eq:mIi_clique}
\Ms{\{i:j, j:k, i:k\}}{i:j} = 
&\min \{0 ,\; \Ms{j:k}{\{i:j, j:k, i:k\}} \;+\; \Ms{i:k}{\{i:j, j:k, i:k\}}\} - \\
&\min \{0 ,\; \Ms{j:k}{\{i:j, j:k, i:k\}} ,\; \Ms{i:k}{\{i:j, j:k, i:k\}}\}\notag
\end{align}

\begin{figure} 
\hbox{
\includegraphics[width=.5\textwidth,height=.8in]{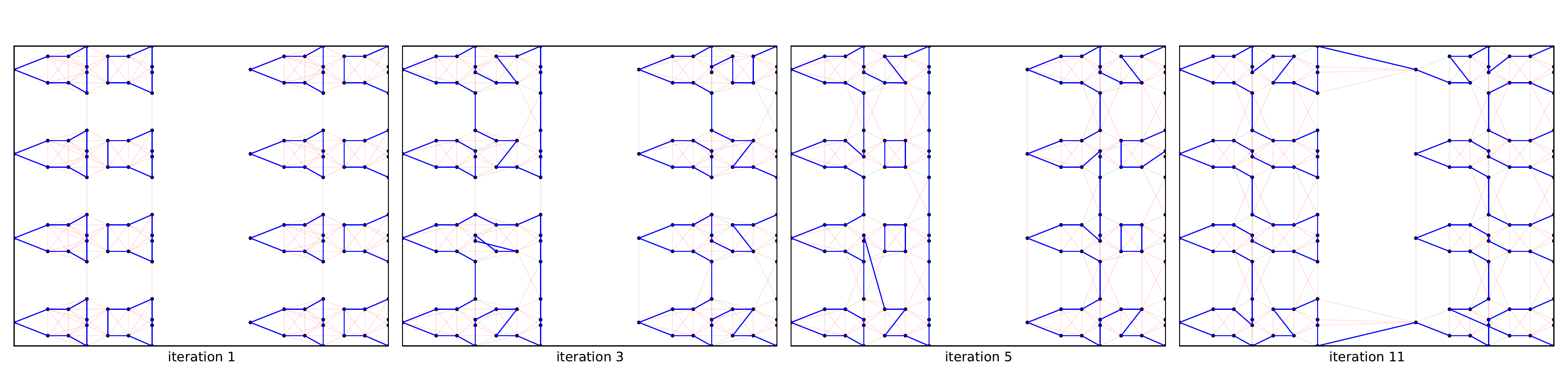}
\hspace{-.01\textwidth}
\includegraphics[width=.25\textwidth,height=.8in]{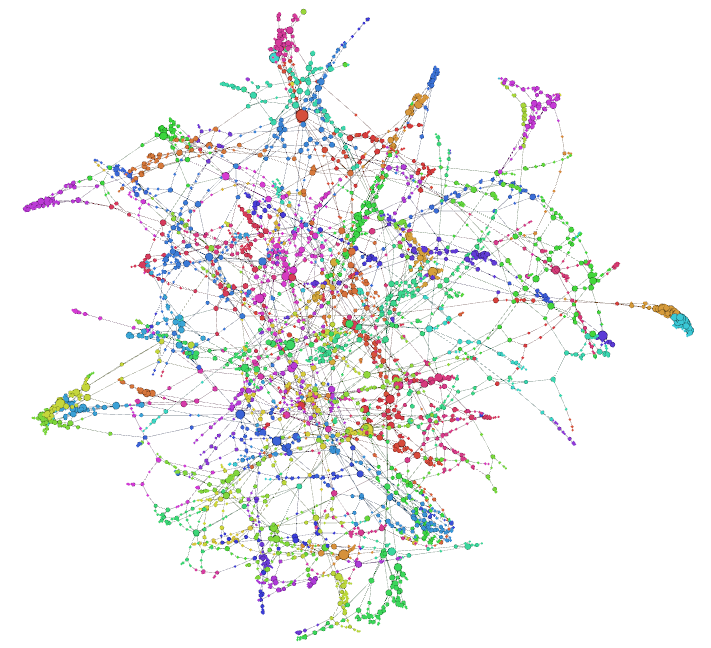}
\hspace{-.01\textwidth}
\includegraphics[width=.25\textwidth,height=.8in]{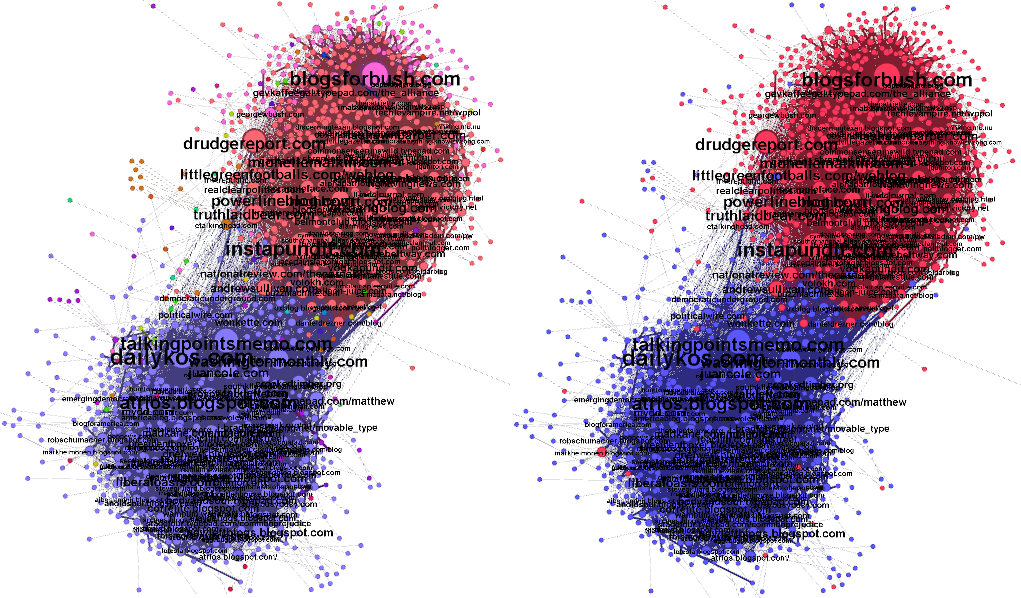}
}
\vspace{-1\in}
\caption{
(left) The message passing results after each augmentation step for the complete graph of printing board instance from \refCitep{tsplib}. 
The blue lines in each figure show the selected edges at the end of message passing. 
The pale red lines show the edges with the bias that,
although negative ($\p_e < 0$), were close to zero.
(middle) Clustering of power network ($N = 4941$) by message passing. Different clusters have different colors and the nodes are scaled by their degree.
(right) Clustering of politician blogs network ($N = 1490$) by message passing and by meta-data -- \ie liberal or conservative.
}
\label{fig:power}
\label{fig:tspexample}
\label{fig:polblogs}
\end{figure}

\subsection{Finding Violations}\label{sec:augmentation}
Due to large number of factors, message passing for the full factor-graph in our applications is not practical. 
Our solution is to start with a minimal set of constraints. 
For TSP, we start with no subtour constraints and for clustering, we start with no clique constraint.
We then use message passing to find marginal beliefs $\p_e$ and  select the edges with positive bias $\p_e  > 0$.

We then find the constraints that are violated. 
For TSP, this is achieved by finding connected components $\set{C} = \{\s_i \subset \V
\}$ 
of the solution  
in $\set{O}(N)$ time and define new subtour constraints for each $\s_i \in \set{C}$ (see \refFigure{fig:tspexample}(left)).

For graph partitioning, we simply look at pairs of positively fixed
edges around each node and if the third edge of the triangle is not positively fixed,
we add the corresponding clique factor to the factor-graph;
see Appendix A for more details.


\begin{figure*}
\includegraphics[width=1\textwidth]{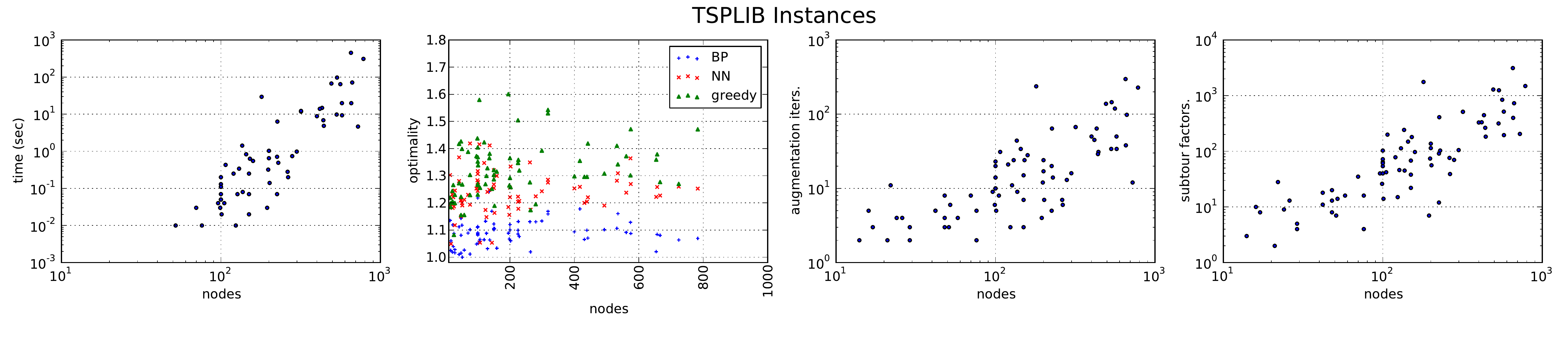}
\includegraphics[width=1\textwidth]{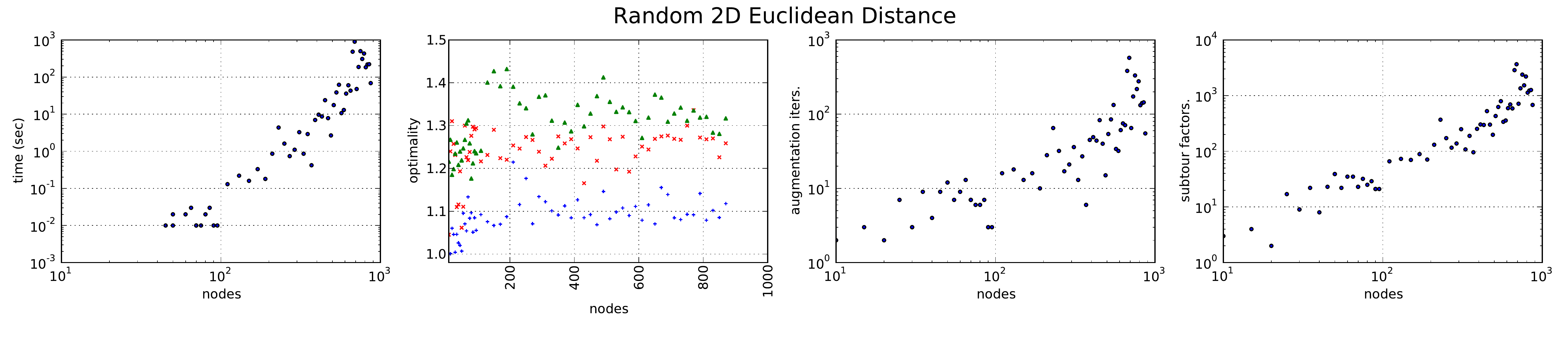}
\includegraphics[width=1\textwidth]{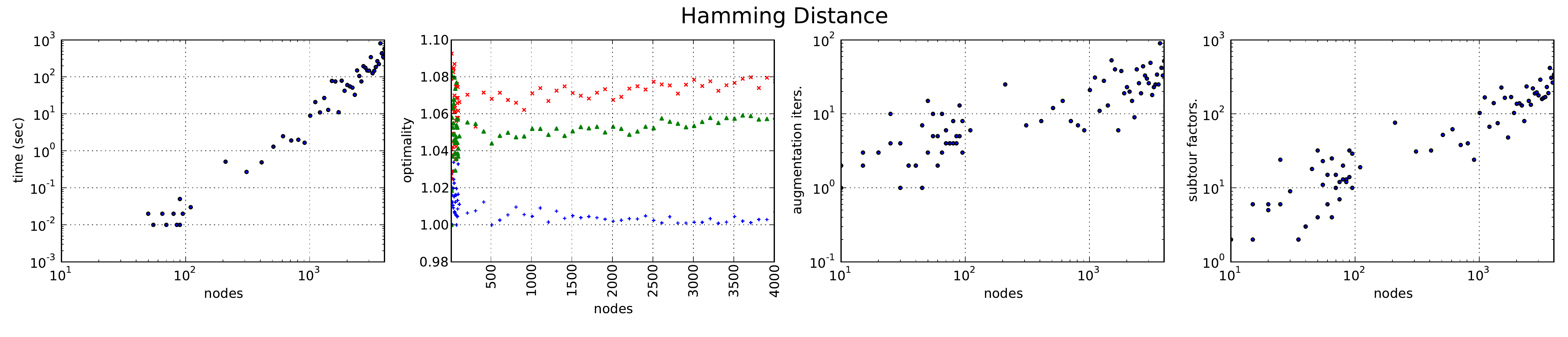}
\includegraphics[width=1\textwidth]{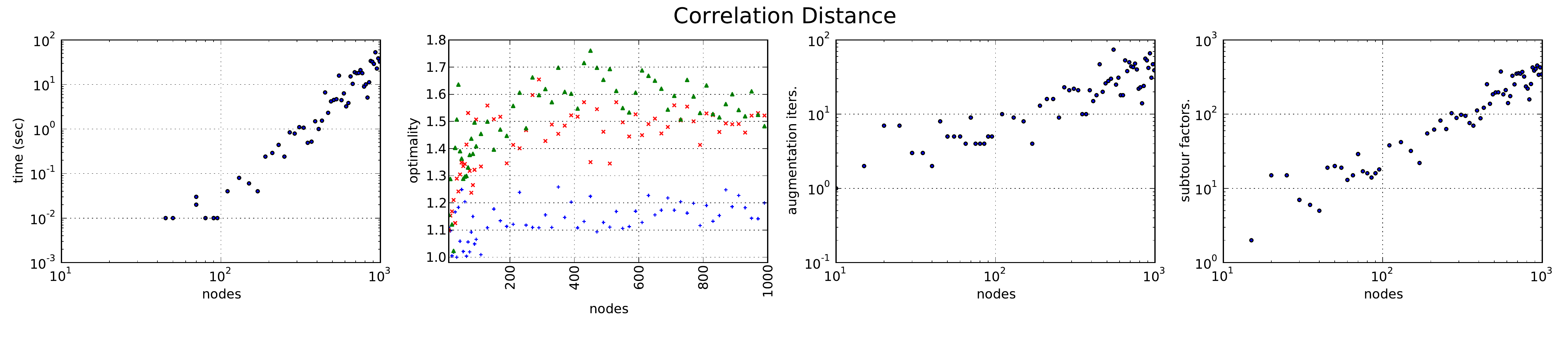}
\includegraphics[width=1\textwidth]{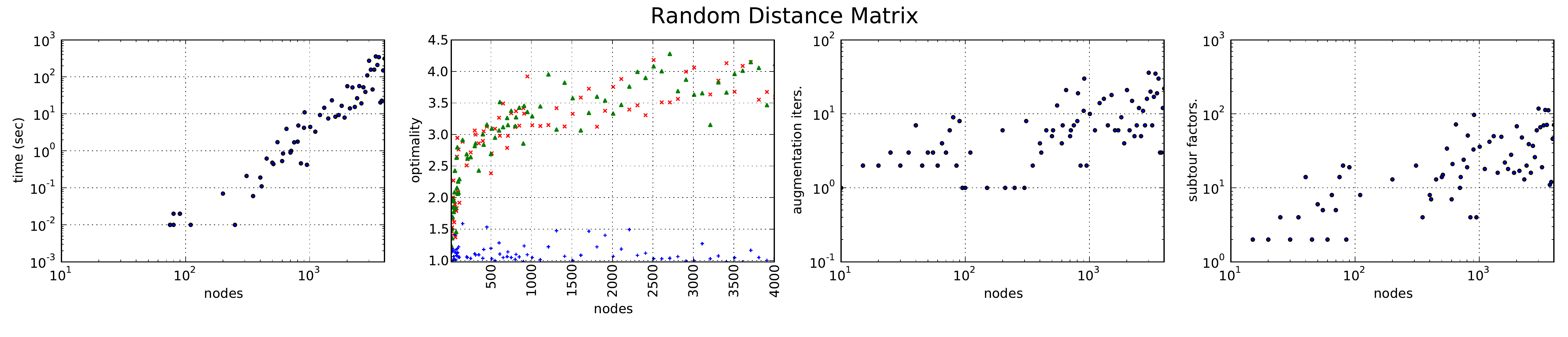}
\caption{\small{Results of message passing for TSP on different benchmark problems.
From left to right, the plots show: (a)~running time, (b)~optimality ratio (compared to Concorde), (c)~iterations of augmentation and 
(d)~number of subtours constraints -- all as a function of number of nodes. 
The optimality is relative to the result reported by Concorde. Note that all plots except optimality are log-log plots where a linear trend shows a monomial relation ($y = a x^{m}$) between the values on the $x$ and $y$ axis, where the slope shows the power $m$.}}
\label{fig:tspresults}
\end{figure*}


\section{Experiments}\label{sec:experiments} 
\vspace*{-1.2em}
\subsection{TSP} 
\vspace*{-1.2em}
Here we evaluate our method over five benchmark datasets:
\textbf{(I)}~TSPLIB, which contains a variety of real-world benchmark instances, the majority of which are 2D or 3D Euclidean or geographic distances.%
\footnote{Geographic distance is the distance on the surface of the earth as a large sphere.}
\textbf{(II)}~Euclidean distance between random points in 2D. 
\textbf{(III)}~Random (symmetric) distance matrices. 
\textbf{(IV)}~Hamming distance between random binary vectors with fixed length (20 bits). This appears in applications such as data compression \refCite{johnson2004compressing} and 
radiation hybrid mapping in genomics~\refCite{ben1997constructing}. 
\textbf{(V)}~Correlation distance between random vectors with 5 random features (\eg using TSP for gene co-clustering \refCite{climer2004take}).
In producing random points and features as well as random distances (in (III)),
we  used uniform distribution over $[0,1]$.

For each of these cases, we report the (a)~run-time, (b)~optimality,
(c)~number of iterations of augmentation and (d)~number of subtour factors at the final iteration.
In all of the experiments, we use Concorde~\refCite{concorde} with its default settings to obtain the optimal solution.%
\footnote{For many larger instances, Concorde (with default setting and using CPLEX as LP solver) was not able to find the optimal solution. 
Nevertheless we used the upper-bound on the optimal produced by Concord in evaluating our method.}
Since there are very large number of TSP solvers, comparison with any particular
method is pointless.
Instead we evaluate the quality of message passing against the ``optimal'' solution.
The results in \refFigure{fig:tspresults}(2nd column from left) 
reports the optimality ratio -- \ie ratio of the tour found by message passing, 
to the optimal tour. 
To demonstrate the non-triviality of these instance, we also report
the optimality ratio for two heuristics that have 
optimality guarantees for metric instances~\refCite{johnson_traveling_1997}:
(a) \textit{nearest neighbour} heuristic ($\set{O}(N^2)$), which incrementally adds the  to any end of the current path the closest city that does not form a loop; 
(b) \textit{greedy} algorithm ($\set{O}(N^2 \log(N))$), which incrementally adds a lowest cost edge to the current edge-set, while avoiding subtours.

In all experiments,
we used the full graph $\G = (\V, \E)$,
which means each iteration of message passing is $\set{O}(N^2 \tau)$, 
where $\tau$ is the number of subtour factors.
All experiments  use $T_{\max} = 200$ iterations, $\epsilon_{\max} =  \hbox{median}\{d(e)\}_{e \in \E}$ and damping with $\lambda = .2$. 
We used decimation, and fixed $10\%$ of the remaining variables (out of $N$) per iteration of decimation.\footnote{Note that here we are only fixing the top $N$ variables with \textit{positive} bias. The remaining $M - N$ variables are automatically clamped to zero.} This increases
the cost of message passing by an $\set{O}(\log(N))$ multiplicative factor, however it often produces better results.

All the plots in \refFigure{fig:tspresults}, except for the second column, are in log-log format.
When using log-log plot, a linear trend shows a monomial relation between $x$ and $y$ axes --
\ie $y = a x ^m$.
Here $m$ indicates the slope of the line in the plot and the intercept corresponds to $\log(a)$. 
By studying the slope of the linear trend in the run-time (left column) in \refFigure{fig:tspresults},
we observe that, for almost all instances,
message passing seems to grow with $N^3$ (\ie slope of $\sim 3$). 
Exceptions are TSPLIB instances,
which seem to pose a greater challenge, and
random distance matrices which 
seem to be easier for message passing.
A similar trend is suggested by the number of subtour constraints and iterations of augmentation, which has a slope of $\sim 1$, suggesting a linear dependence on $N$.
Again the exceptions are TSPLIB instances that grow faster than $N$ and random distance matrices that seem to grow sub-linearly.%
\footnote{Since 
we measured the time in milliseconds,
the first column does not show the instances that had a running time of less than a millisecond.}
Finally, the results in the second column suggests that \emph{message passing is able to find near optimal (in average $\sim 1.1$-optimal) solutions for almost all instances and the quality of tours does not degrade with increasing number of nodes}.

\begin{table*}
  \caption{\small{Comparison of different modularity optimization methods}.}\label{table:clustering}
  \begin{center}
    \scalebox{.5}{
    \begin{tabu}{c | c | c | c |[2pt] r | c | c | l |[2pt] r |c | c | l |[2pt] r| l|[2pt] r| l |[2pt] r| l |[2pt] r| l |}
      \cline{5-20}
      & & & & \multicolumn{4}{ |c |[2pt]}{message passing (full)}&\multicolumn{4}{ |c |[2pt]}{message passing (sparse)}& \multicolumn{2}{ c|[2pt] }{Spin-glass}&\multicolumn{2}{ c|[2pt] }{L-Eigenvector}
      &   \multicolumn{2}{ c|[2pt] }{FastGreedy}&   \multicolumn{2}{ c|[2pt] }{Louvian}\\
      \cline{5-20}
      \begin{sideways}Problem \end{sideways}&
      \begin{sideways} Weighted? \end{sideways}&
     \begin{sideways} Nodes \end{sideways}&
      \begin{sideways} Edges \end{sideways}&
      \begin{sideways} $L$ \end{sideways} &
      \begin{sideways} Cost \end{sideways} &
      \begin{sideways} Modularity \end{sideways} &
      \begin{sideways} Time \end{sideways} &
      \begin{sideways} $L$ \end{sideways} &
      \begin{sideways} Cost \end{sideways} &
      \begin{sideways} Modularity \end{sideways} &
      \begin{sideways} Time \end{sideways} &
      \begin{sideways} Modularity \end{sideways} &
      \begin{sideways} Time \end{sideways} &
      \begin{sideways} Modularity \end{sideways} &
      \begin{sideways} Time \end{sideways} &
      \begin{sideways} Modularity \end{sideways} &
      \begin{sideways} Time \end{sideways} &
 	  \begin{sideways} Modularity \end{sideways} &
      \begin{sideways} Time \end{sideways} 
      \\\tabucline[2pt]{-}

polbooks&y&105&441 
&5461 & 5.68$\%$&0.511&.07
&3624 & 13.55$\%$ &0.506&.04
&0.525&1.648
&0.467&0.179	      
&0.501&0.643
&0.489&0.03 \\

football&y&115&615
&6554&27.85$\%$&0.591&0.41
&5635&17.12$\%$&0.594&0.14
&0.601&0.87
&0.487&0.151
&0.548&0.08
&0.602&0.019 \\
wkarate&n&34&78
&562&12.34$\%$&0.431&0
&431&15.14$\%$&0.401&0
&0.444&0.557
&0.421&0.095
&0.410&0.085
&0.443&0.027 \\
netscience&n&1589&2742
&NA&NA&NA&NA
&53027&\textbf{.0004$\%$}&0.941&2.01
&0.907&8.459
&0.889&0.303
&0.926&0.154
&0.948&0.218 \\
dolphins&y&62&159
&1892&14.02$\%$&0.508&0.01
&1269&6.50$\%$&0.521&0.01
&0.523&0.728
&0.491&0.109
&0.495&0.107
&0.517&0.011 \\
lesmis&n&77&254
&2927&5.14$\%$&0.531&0
&1601&1.7$\%$&0.534&0.01
&0.529&1.31
&0.483&0.081
&0.472&0.073
&0.566&0.011 \\
 celegansneural&n&297&2359
&43957&16.70$\%$&0.391&10.89
&21380&3.16$\%$&0.404&2.82
&0.406&5.849
&0.278&0.188
&0.367&0.12
&0.435&0.031 \\
polblogs&y&1490&19090
&NA&NA&NA&NA
&156753&\textbf{.14$\%$}&0.411&32.75
&0.427&67.674
&0.425&0.33
&0.427&0.305
&0.426&0.099 \\
karate&y&34&78
&562&14.32$\%$&0.355&0
&423&17.54$\%$&0.390&0
&0.417&0.531
&0.393&0.086
&0.380&0.079
&0.395&0.009

        \\\tabucline[2pt]{-}

    \end{tabu}
		}
  \end{center}
\end{table*}

\subsection{Graph Partitioning} 
For graph partitioning, 
we experimented with a set of classic benchmarks%
\footnote{Obtained form Mark Newman's website:
\url{http://www-personal.umich.edu/~mejn/netdata/}
}.
Since the optimization criteria is modularity,
we compared our method only against best known
``modularity optimization'' heuristics: 
(a)~FastModularity\refCite{Clauset05RModularity}, (b)~Louvain~\refCite{blondel2008fast}, (c)~Spin-glass ~\refCite{reichardt2004detecting} and (d)~Leading eigenvector~\refCite{newman2006finding}. 
For message passing, we use $\lambda = .1$, $\epsilon_{\max} =  median\{|\omega(e) - \omega_{null}(e)|\}_{e \in \E \cup \E_{null}}$ and
$T_{\max} = 10$. Here we do not perform any decimation and directly fix the variables based on their bias $\p_e > 0 \Leftrightarrow x_e = 1$. 

\refTable{table:clustering} summarizes our results (see also Figure~\ref{fig:power}(middle,right)). 
Here for each method and each data-set, we report the \textit{time} (in seconds) and the \textit{Modularity} of the communities found by each method.
The table include the results of message passing for both full and sparse null models, 
where we used a constant $\alpha = 20$ to generate our stochastic sparse null model. For message passing, we also included $L = |\E + \E_{null}|$ and the saving in the \textit{cost} using augmentation. This column shows
the percentage of the number of all the constraints considered by the augmentation.
For example, the cost of {\tt .14\% } for the polblogs data-set shows that
augmentation and sparse null model meant using
{\tt .0014} times fewer clique-factors, compared to the full factor-graph.

Overall, the results suggest that our method is comparable to state-of-the-art
in terms both time and quality of clustering. But more importantly it
shows that augmentative message passing is able to find feasible solutions using
a small portion of the constraints.

\section{Conclusion}
We investigate the possibility of using cutting-plane-like, augmentation procedures with
message passing. We used this procedure to solve
two combinatorial problems; TSP and modularity optimization.
In particular, our polynomial-time message passing solution to TSP often finds near-optimal solutions
to a variety of benchmark instances.

Despite losing the guarantees that make cutting plane method very powerful, 
our approach has several 
advantages:
First, message passing is more efficient than LP for structured 
optimization \refCite{yanover2006linear} 
and it is highly parallelizable.
Moreover by directly obtaining integral solutions, it is much easier to find violated constraints.
(Note the cutting plane method for combinatorial problems operates on 
{\em fractional} solutions, whose rounding may eliminate its guarantees.)
For example, for TSPs, our method simply adds violated constraints by finding connected components. 
However, due to non-integral assignments, cutting plane methods require sophisticated tricks to find violations \refCite{applegate2006traveling}.
Although powerful branch-and-cut methods, such as Concorde, are able to exactly solve instances with few thousands of variables, their general run-time on benchmark instances remains exponential~\cite[p495]{concorde}, while our approximation appears to be $\set{O}(N^3)$.
 Overall our studies indicate that augmentative message passing is an efficient procedure for constraint optimization with large number of constraints.


\small{ \bibliography{document} \bibliographystyle{ieeetr}
}

\appendix

\section{Factor-Graphs and PseudoCodes}
\comment{
\begin{figure}
\hbox{
\includegraphics[width=.5\textwidth]{figures/tspfg.pdf}
}
\caption{\small{A simple graph (left), the corresponding binary variable factor-graph for TSP (middle) and its clustering factor-graph (right). 
Circles represent variables. Local factors are grey squares. For TSP, degree constraints are black squares and subtour constraints are white squares. For each subtour constraint, the set $\s$ is the label on the factor. Note that a subtour constraint for $\s$ is also a subtour constraint for $\V \back \s$.
The clustering factor-graph (fully connected null-model) contains one factor, for each triangle.}}\label{fig:fg}
\end{figure}
\refFigure{fig:fg} shows a simple graph and its corresponding TSP and clustering factor-graphs.
}

\begin{algorithm}
\SetKwInOut{Input}{input}\SetKwInOut{Output}{output}
\SetKwFunction{concomp}{ConnectedComponents}
\DontPrintSemicolon
 \Input{Graph $\G = (\V,\E)$, distance function $d: \E \to \Re$, maximum iterations $T_{max}$, damping $\lambda$, threshold $\epsilon_{\max}$.}
 \Output{A subset $\set{T} \subset \E$ of the edges in the tour.}
\DontPrintSemicolon
construct the initial factor-graph\;
initialize the messages $\Ms{i}{e} \leftarrow 0 \;\; \forall i, e \in \partial i$\;
initialize $\p_e \leftarrow d(e)\;\; \forall e \in \E$\;
 \While(\tcp{the augmentation loop}){True}{
  $\epsilon \leftarrow 0$, $T \leftarrow 0$\;
 \While(\tcp{BP loop}){$\epsilon < \epsilon_{\max}$ \textbf{and} $T < T_{\max}$}{
$\epsilon \leftarrow 0$\;
\For(\tcp{inc.\ $\f_{\delta(\s)}$, $\f_{\partial v_i}$}){ \textbf{each} $\f(x_\Is)$}{
  find three lowest values in  $\{\Ms{e}{\Is}\}_{e \in \Is}$\;
  \For{\textbf{each} $e \in \Is$}{
   calculate $\Mst{\Is}{e}$ using eqs~(\ref{eq:mIi_degree},\ref{eq:mIi_subtour})\;
   $\epsilon_{\Is \to e} \leftarrow \Mst{\Is}{e} - \Ms{\Is}{e}$\;
   $\Ms{\Is}{e} \leftarrow \Ms{\Is}{e} + \lambda \epsilon_{\Is \to e}$\;
   $\p_e \leftarrow \p_e + \epsilon_{\Is \to e}$\;
   $\epsilon \leftarrow \max \{ \epsilon, |\epsilon_{\Is \to e}|\}$
   }
 }
 $T \leftarrow T + 1$
}
$\set{T} \leftarrow \{e \in \E \ |\ \p_e > 0 \}$\tcp*{respecting degree constraints.}
$\set{C} \leftarrow$ \concomp{$(\V, \set{T})$}\;
\lIf{$|\set{C}| = 1$}{
  return $\set{T}$
}\lElse{
  augment the factor-graph with $\f_{\s_i}(x_{\delta(\s_i)})\; \forall \s_i \in \set{C}$\;
  initialize $\Ms{\s_i}{e} \leftarrow 0\; \forall \s_i \in \set{C}, e \in \s_i$\;
}
}
\caption{Message Passing for TSP}\label{alg:tsp}
\end{algorithm}

\begin{algorithm}
\SetKwInOut{Input}{input}\SetKwInOut{Output}{output}
\SetKwFunction{concomp}{ConnectedComponents}
\DontPrintSemicolon
 \Input{Graph $\G = (\V,\E)$, weight function $\widetilde{\omega}: \E \to \Re$, maximum iterations $T_{max}$, damping $\lambda$, threshold $\epsilon_{\max}$.}
 \Output{A clustering $\C = \{\C_1,\ldots,\C_K\}$ of nodes.}
construct the null model\;
$\p_{e} \leftarrow 0 \; \forall e \in \E \cup \E_{null}$\;
 \While(\tcp{the augmentation loop}){True}{
  $\epsilon \leftarrow 0$, $T \leftarrow 0$\;
 \While(\tcp{BP loop}){$\epsilon < \epsilon_{\max}$ \textbf{and} $T < T_{\max}$}{
$\epsilon \leftarrow 0$\;
\For{  $e_{i-j} \in \E \cup \E_{null}$}{
  $\p_{e_{i-j}}^{old} \leftarrow \p_{e_{i-j}}$\;
    $\p_{e_{i-j}} \leftarrow \;(\omega(v_i,v_j) - \omega_{null}(v_i, v_j))$\;
    
   \For(\tcp{update beliefs}){  $\Is \ni e_{i-j}$}{
   calculate $\Ms{\Is}{e_{i-j}}$ using \refEq{eq:mIi_clique}\;
   $\p_{e_{i-j}} \leftarrow \p_{e_{i-j}} + \Ms{\Is}{e_{i-j}}$\;
   }
   $\epsilon \leftarrow \max \{ \epsilon, |\p_{e_{i-j}} - \p^{old}_{e_{i-j}} |\}$\;
   \For(\tcp{update msgs.}){  $\Is \ni e_{i-j}$}{
     $\Mst{e_{i-j}}{\Is} \leftarrow \p_{e_{i-j}} - \Ms{\Is}{e_{i-j}}$\;
     $\Ms{e_{i-j}}{\Is} \leftarrow \lambda \Mst{e_{i-j}}{\Is} + (1-\lambda) \Ms{e_{i-j}}{\Is}$
   }
 }
 $T \leftarrow T + 1$
}

\For{  $v_i \in \V$}{
  \For{  $e_{i-j}, e_{i-k} \in \E \cup \E_{null}$}{
    \lIf{$\p_{e_{i-j}} > 0$ \textbf{and} $\p_{e_{i-k}} > 0$ \textbf{and} $\p_{e_{i-k}} \leq 0$}{
      add the corresponding clique factor to the factor-graph\;
}
}
}
\lIf{no factor was added}{\textbf{break} out of the loop}
\lElse{$\Ms{e}{\Is} \leftarrow 0\; \forall \Is, e \in \Is$}
}
$\set{C} \leftarrow$ \concomp{$(\V, \{e \in \E \cup \E_{null} \; |\; \p_{e} > 0 \})$}\;

\caption{Message Passing for Modularity Maximization.}\label{alg:clustering}
\end{algorithm}

Algorithms~\ref{alg:tsp} and \ref{alg:clustering} present the pseudocode for both TSP and graph-partitioning by message passing.
Note that the scheduling of message updates in these two algorithms is very different.
This difference in scheduling is mainly due to the presense of high-order factors in TSP factor-graph and 
intends to minimize the time complexity.
Also, while TSP message passing is re-using the messages from the previous augmentation iteration, 
for clustering, we initialize the messages to zero. This is because the number of factors in each augmentation step for clustering is relatively large and in practice initializing the messages to zero is more efficient. 
For both problems, we have included the message from \textit{local} factors in the marginals and therefore they are ignored during the message update.
In practice we do not need to store \textit{any} of the messages for TSP. Instead we can only keep the three smallest incoming messages to each factor and calculate factor-to-variable messages using these values. The variable-to-factor messages can also be recomputed as required using the marginals and factor-to-variable messages:
$\Ms{e}{\Is}(\x_e)\;=\; \p_e(x_e) - \Ms{\Is}{e}(\x_e)$.

\comment{

\section{Alternative Message Passing Solutions for TSP}
We are aware of two previous attempts at solving TSP by message passing.
\refCite{wangmessage} suggest a binary variable model with $N^2$ variables, in which variable $x_{i,k} = 1$ \textit{iff}
node $i$ is visited at step $k$. However besides being very expensive,
this solution seems to work only for very small instances (Their paper mentions an experiment with $N = 5$).
We previously used a similar formulation with categorical variables that we include here for completeness.

\subsection{Factor-Graph Formulation} \label{sec:formulation}
Let $\x_i \in \set{N} = \{ 1,\ldots, N\}$ denote the time-step in which city $i$ is visited. 
Here $\xv = [\x_1,\ldots,\x_N]$ where $\x_i \neq \x_j \forall i,j$
represents a tour of all cities. 
Note that both the index of a variable $\x_i$ and its value belong to domain $\set{N}$
and they should not be confused.
We define the following distribution over all such tours:
\begin{align} \label{eq:p2}
\p(\xv) \quad = \quad \frac{1}{Z} \prod_{i,j>i} \f_{i,j}(\x_i, \x_j) 
\end{align}
where $Z$ is the normalization constant (\aka partition function) and 
\begin{align}\label{eq:uniform}
\f_{i,j}(\x_i, \x_j) = \left\{ \begin{array}{rl}
0 &\mbox{ if $\x_i = \x_j$ }\\
1 & \mbox{otherwise}
\end{array} \right .
\end{align}

This simple definition of pairwise interaction provides a uniform distribution over all assignments of $\xv$ that are valid orderings --
that is $\p(\xv) = \frac{1}{Z}$ iff $\xv$ is an ordering and $\p(\xv) = 0$ otherwise. 
In this case $Z = N!$ gives the
number of permutations of $N$.

Now consider a more involved case in which different valid orderings assume different 
probabilities. Let $\ddd{i}{j}$ denote the relative preference for city (item) $i$ appearing before city (item) $j$. Here $\ddd{i}{j} = 1$
corresponds to no-preference and $\ddd{i}{j} = 10$ makes this event $10$ times more likely. 
For this define $\f_{i,j}$ as 
\begin{align}\label{eq:ranking}
\f_{i,j}(\x_i, \x_j) =  \left\{ \begin{array}{rl}
0 &\mbox{ if $\x_i = \x_j$}  \\
\ddd{i}{j} &\mbox{ if $\x_i < \x_j$} \\
1 &\mbox{ if $\x_i > \x_j$}
\end{array} \right . 
\end{align}

Once we encode our preferences in the form of local factors, we may use approximate inference to find the \textbf{best ranking} or 
the probability of item $i$ having rank $j$.

We use similar factors to define TSP. 
Let $\dd{i}{j} \in (-\infty, \infty]$ be the distance from the city $i$ to the city $j$ and define  
$\f_{i,j}$ as:
\begin{align}\label{eq:tsp1}
&\f_{i,j}(\x_i, \x_j) = \\ &\left\{ \begin{array}{rl}
e^{-\infty} = 0 &\mbox{ if $\x_i = \x_j$} \notag\\
e^{- \dd{i}{j}} &\mbox{ if $\x_i = \x_j - 1$ or $(i,j) = (1,N)$} \notag\\
e^{- \dd{j}{i}} &\mbox{ if $\x_i = \x_j + 1$ or $(i,j) = (N,1)$} \notag\\
e^{0} = 1 & \mbox{otherwise} 
\end{array} \right .\notag
\end{align}
where $\f_{i,j}$ only depends on the distances $\dd{i}{j}$ and $\dd{j}{i}$.
Also from now on we assume modular arithmetic --\ie $N + 1 \equiv 0$ and $0 -1 \equiv N$.
This is to indicate that the first city is visited after the last one, forming a loop.

\begin{figure}
\[ \left[ \scalemath{.7}{ \begin{array}{ccccccc}
0 & e^{-\dd{i}{j}} & 1  & \cdots  &1 &1 & e^{-\dd{j}{i}}\\
e^{-\dd{j}{i}} & 0 & e^{-\dd{i}{j}} & \cdots & 1 & 1 & 1\\
1 & e^{-\dd{j}{i}} & 0 & \cdots  & 1 & 1 & 1\\
\vdots & \vdots & \vdots &\ddots  & \vdots & \vdots & \vdots \\
1 & 1 & 1 & \cdots & 0 & e^{-\dd{i}{j}} & 1\\
1 & 1 & 1& \cdots& e^{-\dd{j}{i}} & 0 & e^{-\dd{i}{j}}\\
e^{-\dd{i}{j}} & 1  & 1 &\cdots & 1 & e^{-\dd{j}{i}} & 0\\
\end{array} } \right] \]
\caption{The tabular form of $\f_{i,j}(\x_i, \x_j)$ as defined in \refEq{eq:tsp1}.}
\label{fig:tsp1}
\end{figure} 

\refFigure{fig:tsp1} shows the tabular form of this factor. Zero diagonal of $\f_{i,j}$ 
ensures $\p(\xv) = 0$ whenever $\x_i = \x_j$, 
and two other bands correspond to the cases where $\x_i$ and $\x_j$ are visited one after the other.
More specifically, $\x_i$ changes along the rows and $\x_j$ along the columns. 
The band below diagonal corresponds to $\x_i = \x_j + 1$ and the upper band corresponds to $\x_i = \x_j - 1$. 
If these two cities are not visited consecutively then the value of $1$ 
ensures that this factor has no contribution to $\p(\xv)$.
Here the optimal tour $\xv^*$ corresponds to the Maximum A Posteriori (MAP) assignment of $\p$:
\begin{align*}
\xv^* =& \arg_{\xv} \max \p(\xv) \\
=& \arg_{\xv} \max \sum_{i,j>i} \log(\f_{i,j}(\x_i,\x_j)) \\
 =& \arg_{\xv, \x_i \neq \x_j} \min \sum_{i, j >i} \bigg ( \sum_{t=1}^{N}  \delta(\x_i, t) \delta(\x_j, t+1) \dd{i}{j} \bigg )
\end{align*}
 In the last equation we have expanded $\log(\f_{i,j}(\x_i, \x_j))$ in the form of a summation over distances.
 Here $\delta(\x_i, t) = 1$ iff $\x_i = t$ and $0$ otherwise. This indicates that the cost of a path is equal to the
 negative log-probability of assignment $\xv$.
 
\subsubsection{Variations of TSP and VRP}
 
  When $\dd{i}{j} \neq \dd{j}{i}$ for some $i$ and $j$ the problem is known as  \textbf{asymmetric TSP}. 
  
  When dealing with a graph it is enough to use the same definition of $\f_{i,j}$ and
 set $\dd{i}{j} = \dd{j}{i} = 0$ iff two vertices are adjacent and $\infty$ otherwise.
 Here the MAP assignment is a \textbf{Hamiltonian cycle} and the partition 
 function $Z$ is equal to the number of Hamiltonian cycles in the graph.

 \begin{figure}
\[ \left[ \scalemath{.7}{\begin{array}{ccccccccc}
0 & 0 & 0 & \cdots & 0 & 0&0\\
e^{-\dd{j}{i}} & 0 & 0 & \cdots & 0 & 0 &0\\
1 & e^{-\dd{j}{i}} & 0 & \cdots & 0 & 0 &0\\
\vdots & \vdots & \vdots & \ddots & \vdots &\vdots & \vdots\\
1 & 1 & 1 & \cdots &e^{-\dd{j}{i}} & 0 & 0\\
1 & 1 & 1 &\cdots &1 & e^{-\dd{j}{i}} & 0\\
\end{array}} \right]\] 
\caption{The tabular form of $\f_{i,j}(\x_i, \x_j)$ used in sequential ordering problem (\refEq{eq:sop}).}
\label{fig:sop}
\end{figure}

 \textbf{Sequential Ordering Problem} seeks the minimum cost assignment where some nodes have to precede others
 in a path that starts at the first city --\ie $\x_i = 1$. 
 Note that precedence relationship is transitive-- \ie if $\x_i > \x_j$ and $\x_j > \x_k$ then $\x_i > \x_k$. 
 However if we define a set of precedence constraints as follows, then the transitivity relation is automatically enforced.
 Here for $\x_i > \x_j$ we define $\f_{i,j}$ as:
\begin{align}\label{eq:sop}
\f_{i,j}(\x_i, \x_j) = \left\{ \begin{array}{rl}
e^{- \dd{j}{i}} &\mbox{ if $\x_i = \x_j + 1$ } \\
e^{- \infty} = 0 & \mbox{ if $\x_i \leq x_j$} \\
e^{0} = 1 & \mbox{otherwise}
\end{array} \right .
\end{align}
where $\f_{i,j}$ in tabular form takes a lower triangular form (see \refFigure{fig:sop}).

A local factor $\f_i(\x_i)$ can constrain or penalize certain visiting time-steps $\set{T}_i \subseteq \set{N}$ for city $i$. For example $\f_i(\x_i = \ell) = 0$ prevents visiting city $i$
at time-step $\ell$.  
  
  A more involved case penalizes the \textbf{relative visiting time} of two cities. Sequential ordering constraint is a special
  case of this type. Penalty of this type can be specified by defining the factor $\f_{i,j}$
   for cities $i$ and $j$ as
\begin{align}\label{eq:timewindow}
 & \f_{i,j}(\x_i, \x_j) = & & \\ &\left\{ \begin{array}{rl}
  e^{- \dd{i}{j}} &\mbox{ if $\x_i = \x_j - 1$ or $(i,j) = (1,N)$} \notag\\
  e^{- \dd{j}{i}} &\mbox{ if $\x_i = \x_j + 1$ or $(i,j) = (N,1)$} \notag\\
  e^{- \infty} &\mbox{ if $\x_i = \x_j$} \notag\\
  e^{- \alpha_{i > j}} & \mbox{ if $1 < \x_i - \x_j < T$ or $1 < \x_j + (T - \x_i) \leq T$} \notag\\
  e^{-\alpha_{i < j}} & \mbox{ if $1 < \x_j - \x_i < T$ or $1 < \x_i + (T - \x_j) \leq T$} \notag\\
  e^{-\beta} & \mbox{otherwise} 
\end{array} \right .\notag
\end{align}
where $\alpha_{i > j}$ ($\alpha_{i < j}$) is the penalty for visiting city $i$ ($j$) within $T$ steps after visiting $j$ ($i$). 
For example $\beta = \alpha_{i < j} = \infty$ and $\alpha_{i > j} = 0$
only allows the cases in which, city $i$ is visited  within a time-window of length $T$ after city $j$ (see \refFigure{fig:timewindow} for tabular form). 
A different constraint may require two cities to be visited \emph{after} $T$ time-steps. 
Setting $\beta = 0$ and $\alpha_{i>j}=\alpha_{i < j} = \dd{i}{j} = \dd{j}{i} = \infty$ enforces this.
In the factor above to avoid contradictions where 4th and 5th condition in \refEq{eq:timewindow} 
are both satisfied, we assume $T < \frac{N-1}{2}$.
Other cases in which $T > \frac{N-1}{2}$ are plausible if we remove either one of lines (4) or (5)
in \refEq{eq:timewindow}
 
 \begin{figure}
\[ \left[ \scalemath{.7}{ \begin{array}{ccccccc}
0 & e^{-\dd{i}{j}} & e^{-\alpha_{i < j}} & \cdots & e^{-\beta} & e^{-\alpha_{i > j}} & e^{-\dd{j}{i}}\\
e^{-\dd{j}{i}} & 0 & e^{-\dd{i}{j}} &\cdots & e^{-\beta} & e^{-\beta} & e^{-\alpha_{i > j}}\\
e^{-\alpha_{i > j}} & e^{-\dd{j}{i}} & 0 & \cdots & e^{-\beta} & e^{-\beta} & e^{-\beta}\\
\vdots & \vdots & \vdots  & \ddots & \vdots & \vdots & \vdots\\
e^{-\beta} & e^{-\beta} & e^{-\beta} &\cdots & 0 & e^{-\dd{i}{j}} & e^{-\alpha_{i < j}}\\
e^{-\alpha_{i < j}} & e^{-\beta} & e^{-\beta} &\cdots & e^{-\dd{j}{i}} & 0 & e^{-\dd{i}{j}}\\
e^{-\dd{i}{j}} & e^{-\alpha_{i < j}} &e^{-\beta} &  \cdots &  e^{-\alpha_{i > j}} & e^{-\dd{j}{i}} & 0\\
\end{array} } \right] \]
\caption{The tabular form of time-window constrained $\f_{i,j}$ of \refEq{eq:timewindow} for $T =2$.}
\label{fig:timewindow}
\end{figure} 
  
 \subsubsection{Vehicle Routing Problem}
 TSP can been seen as a special form of VRP with a single vehicle. 
 Here we use a trick similar to the one used by \refCitep{lenstra_simple_1975} to go in the opposite direction -- \ie reduce VRP to TSP.
 In this problem the depot is always the first node to visit, therefore we need to have $N$ variables -- one for each customer.
 Similar to TSP the value of each variable is the time-step in which the customer is visited.
 Let $M$ denote the number of vehicles in a VRP instance. We reduce the VRP to TSP by 
  defining the distances from $M$ depot replicates to each of $N$ cities identical to that 
 of the first depot. Let $\set{N} = \{ 1,\ldots, N\}$ be the set of customers indices and $\set{M} = \{N+1,\ldots, N+M\}$ 
 denote the set of depot indices. We define distances in the new TSP-reduction as follows: 
\begin{align}
\dd{i}{j}' = \left\{ \begin{array}{rl}
\dd{i}{j} &\mbox{ if $ i,j \in \set{N}$ }\\
\dd{i}{\text{depot}} &\mbox{ if $i\in \set{N}$ and $j \in \set{M}$ } \notag \\
\dd{\text{depot}}{j} &\mbox{ if $j\in \set{N}$ and $i \in \set{M}$ } \notag\\
\gamma & \mbox{ if $ i,j \in \set{M}$} \notag\\
\end{array} \right .\notag
\end{align}

This implies that each depot-replicate is a city that should be visited in the tour. 
Setting $\gamma = \infty$ means that depots will not be visited consecutively, which means we are producing \emph{exactly} $M$ loops that
start and end at the depot. Setting $\gamma = 0$, allows for consecutive visits to depots at no cost. This corresponds to setting an upper-bound
on the number of vehicles in VRP.

We consider a limited form of \textbf{capacitated VRP} 
in which each vehicle has the capacity to deliver to at maximum $N_m$ customers.
This can be enforced by setting time-constraints on relative visiting time of each of depot replicates.
Let $m_1,\ldots, m_{\ell},\ldots,m_M$ be an arbitrary ordering of $M$ depot-replicates. For each $\x_{m_{\ell}}$ 
 we add the following time-window constraints: 
 $\x_{m_{\ell - 1}} <\x_{m_{\ell}} < \x_{m_{\ell - 1}+N_{m_\ell}}$ -- \ie
 the next depot must be visited at most $N_{m_{\ell}}$ time-steps
 after the current depot.

\subsection{Estimating the MAP Assignment}\label{sec:inference}
In this section we first explain the common approach to finding a MAP assignment in a PGM -- \ie using max-product BP.
Interestingly we observe a better performance using sum-product BP and decimation procedure. 
In both cases, it is possible to pass BP messages through $\f_{i,j}(\x_i, \x_j)$ efficiently (in $\set{O}(N)$).
We further discuss efficient sum-product message passing in \refSection{sec:efficient}.   

\subsection{First Attempt: Max-Product BP}
BP is an inference procedure that involves a set of recursive update equations
for a set of messages ($\Ms{i}{j}$ and $\Ms{j}{i}$) sent between 
each pair of connected variables $\x_i$ and $\x_j$. 
Max-product BP is a variation of BP that is commonly used to find MAP assignment:
\begin{align}
\Ms{i}{j}(\x_j)  \propto \max\{ \f_i(\x_i)\; \f_{i,j}(\x_i, \x_j) \prod_{k \in \set{N} \back i,j} \Ms{k}{i}(\x_i)\}_{\x_i}
\end{align}
where $\set{N} \back i,j$ is the set of all variable indices minus $i$ and $j$.
Here the message sent from node $i$ to $j$ is proportional to
 the product of factors $\f_i$, $\f_{i,j}$ and all the incoming messages
to node $i$ except for the message from $j$. 
This function is then maximized over all values of $\x_i$ to produce the message $\Ms{i}{j}$ that
goes from $i$ to $j$.
This recursive update procedure is applied for all $i$ and $j$, 
repeatedly until a fixed point is reached. 
After convergence the max-marginal over each variable is calculated as the product of all incoming messages and the local factor
\begin{align}\label{eq:marg}
\ph(\x_i) \propto \f_i(\x_i) \prod_{j \in \set{N}\back i}\Ms{j}{i}(\x_i)
\end{align}

Then each variable is assigned to its most probable value 
$\xh_i^* = \arg_{\x_i} \max \f(\x_i)$ independently, 
producing an estimate $\xvh^*$ of MAP assignment $\xv^*$.
Usually a decimation procedure is employed for tie-breaking.

We observe that even if converged, max-product BP often returned low quality results 
in this factor-graph formulation of TSP.
We therefore instead use sum-product BP with decimation to produce better MAP estimates.
 
\subsection{Sum-Product BP and Decimation}

Sum-product BP provides an estimate of marginals $\p(\x_i) = \sum_{\xv \back \x_i} \p(\xv)$ using the following recursive message update:
\begin{align}
\Ms{i}{j}(\x_j) \;  \propto \; \sum_{\x_i} \; \f_i(\x_i)\; \f_{i,j}(\x_i, \x_j) \prod_{k \in \set{N} \back i,j} \Ms{k}{i}(\x_i) \label{eq:bp}
\end{align}

If convergent the estimate of marginals is given by \refEq{eq:marg}, 
where we can use $\xh_i^* = \arg_{\x_i} \max \ph(\x_i)$
as an estimate of the max-marginal and $\xvh^* = [\xh_1^*,\ldots,\xh_N^*]$ as a MAP estimate.
The logic behind this becomes more clear when we introduce a temperature parameter.
 
\subsubsection{Temperature}
Distribution $\p(\xv)$ \refEq{eq:p2} changes dramatically if we multiply distance between cities
 by the same constant. However obviously
the optimal solution does not change. This variability in $\p(\xv)$ is modeled by a Temperature parameter $\tau$:
\begin{align} \label{eq:pt}
\p(\xv,\tau) \quad = \quad \frac{1}{Z_\tau} \prod_{i} \f_i(\x_i)^{\frac{1}{\tau}} \prod_{j>i} \f_{i,j}^{\frac{1}{\tau}}(\x_i, \x_j) 
\end{align}
where $\p(\xv, 1) = \p(\xv)$. By decreasing $\tau$ towards zero, $\p(\xv, \tau)$ becomes more peaked around more probable assignment where at the limit
$\lim_{\tau \to 0} \p(\x, \tau)$ only the MAP assignment(s) has non-zero probability.
This suggests that we can use the maximum of sum-product marginals at low temperatures to estimate MAP assignment (\refCite{weiss_map_2007}).

However there is a drawback to applying BP at lower temperatures; in many problems 
lower temperatures correspond to a more difficult 
inference problem which degrades BP's performance. 
Fortunately we can pick between several BP results at different temperatures,
based on the quality of their MAP estimates. To this end we first normalize all distances $\dd{i}{j}$ by dividing by maximum distance, between
any two cities. Then we run BP at few temperatures $\tau \in \{.1, .2,.5,1,2\}$ and pick the best result.

\begin{figure}
 \includegraphics[width=0.25\textwidth]{figures/hamiltonian.pdf}
 \caption{The graph of two platonic solids; octahedron with 32 and cube with 12 Hamiltonian cycles. }
 \label{fig:hamiltonian}
 \end{figure}

\begin{table}
\caption{Probability of visiting each node (row) at each time-step (col) 
in a Hamiltonian cycles of octahedron. The estimates are using Junction Tree (JT)
and BP.}
\label{table:hamiltonian}
\scalebox{.7}{ 
\begin{tabular}{|r |  c  c  c c c c|}
\hline
node & $\p(1)$& $\p(2)$&$\p(3)$&$\p(4)$&$\p(6)$&$\p(7)$ \\
\hline
JT&&&&&&\\
1 & 1&0&0&0&0&0 \\
2,\ldots,5 & 0&.25&.187&.125&.187&.25\\
6 & 0&0&.25&.5&.25&0\\
\hline 
BP&&&&&&\\
1 & 1&0&0&0&0&0 \\
2,\ldots,5 & 0 &.256&.160&.165&.160&.256\\
6 & 0&0&.335&.328&.335&0\\
\hline
\end{tabular}
}
\end{table}

\subsubsection{Decimation}
Another issue with sum-product BP to estimate the MAP assignment (at $\tau > 0$) 
is that $\arg_{\x_i} \max \ph(\x_i)$ for variables $\x_i$ and $\x_j$ may correspond to distinct high probability assignments -- 
that is while both $\xv$ and $\xv'$ have high probabilities, setting $\x_i$ according to one, and $\x_j$ according to the other,
may correspond to a low probability assignment.

To solve this issue, we use a decimation procedure. In BP-guided decimation, after each application of BP
a single most biased variable (or a small subset $\set{B}$) is clamped to its most probable value. Then BP is ran again on the clamped PGM.
This procedure is repeated until all variables are fixed. 



\begin{table}
\caption{symmetric/asymmetric TSP and sequential ordering problem instances from TSPLIB }
\label{table:tsplib}
\scalebox{.7}{ 
\begin{tabular}{|r | c  | c | c | l|}
\hline
  name & $N$& \# const  &BP &opt \\
  \hline                        
  sym\\
  \hline
  att48 &  48 && 11785 & 10628\\
  bayg29 &  29 && 1751& 1610\\
  bays29 &  29 && 2129& 2020\\
  berlin52 & 52 && 8310& 7542\\
  brazil58 & 58 && 28434& 25395\\
  burma14 & 14 && 3599 & 3323\\
  dantzig42 & 42 && 729 & 699\\
  eli51& 51 && 467 & 426\\
  eli76  & 76 && 604 & 538\\
  fri26& 26 && 1032 & 937\\
  gr17 & 17 && 2095 & 2085\\
  gr21 & 21 &&2932 & 2707\\
  gr24 & 24 && 1346 & 1272\\
  gr48 & 48 && 5389& 5046\\
  gr96 & 96 && 55432 & 55209\\
  hk48 & 48 && 12940& 11461\\
  kroA100 & 100&& 24841 & 21282 \\
  kroB100&  100 && 25580 & 22141\\
  kroC100&  100 && 23743 & 20749\\
  kroD100&  100 && 24042 & 21294\\
  kroE100&  100 && 23953 & 22068\\
  pr76 &  76 && 125183& 108159\\
  rat99& 99 && 1338 & 1211\\
  rd100&  100 && 9370& 7910\\
  st70&  70 && 724 & 675\\
  swiss42 & 42 && 1417& 1273 \\
  ulysses16 &16 && 7234& 6928\\
  ulysses22 &22 && 7856& 7132\\
  \hline
  asym\\
  \hline
    br17 & 17 && 41 & 39\\
  ft53 &  53 && 8151& 6905\\
  ft70 &  70 && 41868& 38673\\
  ftv33&  34 && 1524 & 1276\\
  ftv35&  36 && 1671& 1473\\
  ftv38&  39 && 1761& 1530\\
  ftv44&  45 && 2078& 1613\\
  ftv47&  48 && 2065& 1776\\
  ftv55&  56 && 2245& 1608\\
  ftv64&  65 && 2346& 1839\\
  ftv70&  71 && 2599& 1950\\
  p43 &  43 && 5959& 5620\\
  ry48p&  48 && 15672& 14422\\
  \hline  
  sop\\
  \hline
   ESC07&  9 & 6 &2550 & 2125\\
  ESC11&  13 & 3&2423& 2075\\
  ESC12&  14 & 7&1802& 1675\\
  ESC25&  27 & 9&2884& 1681\\
  ECS47&  49 & 10&3472&1288 \\
  ESC63&  65 & 95 &75 &62 \\
  ESC78&  80 & 77 &20440&18230 \\
  br17.10&  18&10 & 102 &55 \\
  br17.12&  18&12 & 75 & 55\\
  ft53.1&  54 & 12&10327&7570 \\
  ft53.2&  54 & 25&12739&8335 \\
  ft53.3&  54 & 48&17073 &10935 \\
  ft53.4&  54 & 63&19541& 14425\\
  ft70.1&  71 & 17&46750 & 39313\\
  ft70.2&  71 & 35&48344&41778 \\
  ft70.3&  71 & 68&56005&44732\\
  ft70.4&  71 & 86&64396&53882\\
  kro124p.1&  101 & 25&61228&42845\\
  kro124p.2&  101 & 49&68133&45848 \\
  kro124p.3&  101 & 97&85470&55649\\
  kro124p.4&  101 & 131&123548&80753 \\
  p43.1&  44 & 9&30640& 27990\\
  p43.2&  44 & 20&57280& 28330\\
  p434.4&  44 & 50&85980& 82960\\
  prob42 &  42 &10 & 450 & 243\\
  rbg050c&  52 & 256&579 &467\\
  ry48p.1 &  49 &11 &21428 & 15935\\
  ry48p.2 &  49 &23 &23232& 17071\\
  ry48p.3& 49 &42 &29243 & 20051\\
  ry48p.4& 49 &58 &37037 & 31446\\
  \hline  
  \end{tabular}
  }
\end{table}   

\subsubsection{Counting Hamiltonian Cycles}
When the distances are either $0$ or $\infty$, the partition function of \refEq{eq:p2}
counts the number of distinct Hamiltonian cycles. Here using Junction Tree method 
(\refCite{lauritzen_local_1988}) 
we found the correct number of unique cycles 
for two platonic shapes of \refFigure{fig:hamiltonian}. \refTable{table:hamiltonian} shows the probability of
visiting each of $6$ nodes of octahedron at any time-step during the cycle that starts at node $1$.

\subsection{Efficient Message Passing}\label{sec:efficient}
Naive calculation of message update of \refEq{eq:bp} requires $\set{O}(N^2)$ operations. Since there are 
$\set{O}(N^2)$ of such messages, each iteration of BP is in the order of $\set{O}(N^4)$.

 In this section we show that the factors we used in different variations of TSP and VRP allow efficient update in $\set{O}(N^3)$ 
 instead of $\set{O}(N^4)$.\footnote{This is assuming no decimation.}
 
In order to write simplified BP update for $\Ms{i}{j}$, let $\Ms{\back j}{i}$ denote
\begin{align*}
\Ms{\back j}{i}(\x_i) \quad = \quad \frac{1}{Z_{\back j \to i}} \;\phi_i(\x_i)\; \prod_{k \in \set{N} \back i,j} \Ms{k}{i}(\x_i)
\end{align*}
where $Z_{\back j \to i}$ is a normalization constant. This simplifies the update of \refEq{eq:bp}
\begin{align*}
\Ms{i}{j}(\x_j) \quad \propto \quad \sum_{\x_i} \Ms{\back j}{i}(\x_i)\; \f_{i,j}(\x_i , \x_j) 
\end{align*}

Naive calculation of $\Ms{\back j}{i}$ requires $\set{O}(N^2)$ operations. 
However it can be obtained from $\ph(\x_i)$ of \refEq{eq:marg} in linear time,
resulting in total of $\set{O}(N^3)$ complexity.\footnote{Here we calculate $N$ vectors of $\ph(\x_i)$ each in $\set{O}(N^2)$. 
To avoid degeneracies and division by zero we need to replace zero probabilities 
in messages with small values.} 
 Here we show how to calculate $\Ms{j}{i}$ 
given $\Ms{\back j}{i}$ for different type of factors in linear time.
 We start by the simple TSP factor of \refEq{eq:tsp1}. The tabular form of this factor is shown in \refFigure{fig:tsp1}.
 Calculating $\Ms{i}{j}$ corresponds to multiplication of the rows by $\Ms{\back j}{i}$ and summing over columns.
  
Simple calculation shows\footnote{As before, we define $x_\ell-1$ for $x_\ell = 0$, equal to $N$ and $x_\ell + 1$ for $x_\ell = N$ equal to $0$.}
 \begin{align*}
 \Ms{i}{j}(\x_j)\; & \propto \; Z_{\back j \to i} - \Ms{\back j}{i}(\x_i) \\
 +& (1 - \Ms{\back j}{i}(\x_i - 1))e^{-\dd{j}{i}} \notag\\
 +& (1 - \Ms{\back j}{i}(\x_i +1))e^{-\dd{i}{j}}  \notag
 \end{align*}
 Since we need to calculate $Z_{\back j \to i}$ only once, this message update for $1 \leq \x_j \leq N$ is linear in $N$.

We now consider the factor used by sequential ordering problem. The tabular form of \refEq{eq:sop} is shown in \refFigure{fig:sop}.
In this case we can calculate the messages incrementally, starting from $\Ms{i}{j}(N) = 0$ and $\Ms{i}{j}(N-1) = \Ms{\back j}{i}(N-1)e^{- \dd{j}{i}} $,
 we may calculate $\Ms{i}{j}(\x_j - 1)$
from $\Ms{i}{j}(\x_j)$:
  \begin{align*}
  \Ms{i}{j}(\x_j - 1) \; &\propto \; \Ms{i}{j}(\x_j) + \\ 
  &e^{- \dd{j}{i}} \Ms{\back j}{i}(\x_j -2) + \notag\\
  & (1 - e^{- \dd{j}{i}})\Ms{\back j}{i}(\x_j-1)  \notag
  \end{align*}

The message in the other direction (\ie $\Ms{j}{i}$) is calculated slightly differently but uses the same trick.
 
Finally consider the more general factor form of \refEq{eq:timewindow}.
Looking at the tabular form of \refFigure{fig:timewindow}, it is feasible to first 
calculate $\Ms{i}{j}(1)$ in $\set{O}(N)$ and then
recursively calculate $\Ms{i}{j}(\x_j+1)$ given $\Ms{i}{j}(\x_j)$. Since only
few values in the summation change from $\x_j$ to $\x_j + 1$, each recursive calculation can be
performed in $\set{O}(1)$. The same trick is applicable to the ranking factor of \refEq{eq:ranking}.

\begin{figure}
 \includegraphics[width=0.4\textwidth]{figures/randmat.pdf}
 \caption{Comparison of tour length on random distance matrices of various size.}
 \label{fig:random}
 \end{figure}

\subsection{Experiments}\label{sec:experimentss}
We experimented with various instances (with $N \leq 100$) from TSPLIB.
as well as random distance matrix instances, where $\dd{i}{j} \sim \mathbb{U}(0,1)$.
Here BP-guided decimation was initially ran for maximum of 
$50$ iterations and after each iteration of decimation, five BP iteration was performed.

}
\end{document}